\documentclass[10pt,twocolumn,letterpaper]{article}
\usepackage[pagenumbers]{cvpr}
\usepackage[T1]{fontenc}

\DeclareTextFontCommand{\texttt}{\fontencoding{OT1}\ttfamily}
\usepackage{amsmath,amssymb,booktabs,array,tabularx}
\usepackage{xcolor}
\usepackage{colortbl}
\definecolor{reportblue}{RGB}{38,100,151}
\usepackage[breaklinks,colorlinks,allcolors=reportblue]{hyperref}
\hypersetup{pdftitle={How Far Can GPT-6-Astra Go? Evaluating capabilities in Zero-Shot Vision-and-Language Navigation},pdfauthor={Guangzhao Dai, Qi Wu, and Bin Zhu}}
\newcolumntype{Y}{>{\raggedright\arraybackslash}X}

\graphicspath{{figures/r1_report_20260915/}}
\title{How Far Can GPT-6-Astra Go?\\[3pt]
{\large Evaluating capabilities in Zero-Shot Vision-and-Language Navigation}}
\author{Guangzhao Dai$^1$,
Qi Wu$^2$,
Bin Zhu$^{1\dagger}$\\
$^1$School of Computing and Information Systems, Singapore Management University \quad \\
$^2$Australia Institute for Machine Learning\\
\normalsize{$^\dagger$Corresponding author and project lead}\\[3pt]
\normalsize{Website: \url{https://daiguangzhao.github.io/gpt-6-astra-vln/}}
}

\makeatletter
\patchcmd{\@maketitle}{\vskip .375in}{\vskip 0pt}{}{}
\patchcmd{\@maketitle}{\vspace*{12pt}}{\vspace*{4pt}}{}{}
\patchcmd{\@maketitle}{\vspace*{24pt}}{\vspace*{12pt}}{}{}
\g@addto@macro\@maketitle{%
  \vspace{-24pt}
  \begin{center}
    \includegraphics[width=\textwidth]{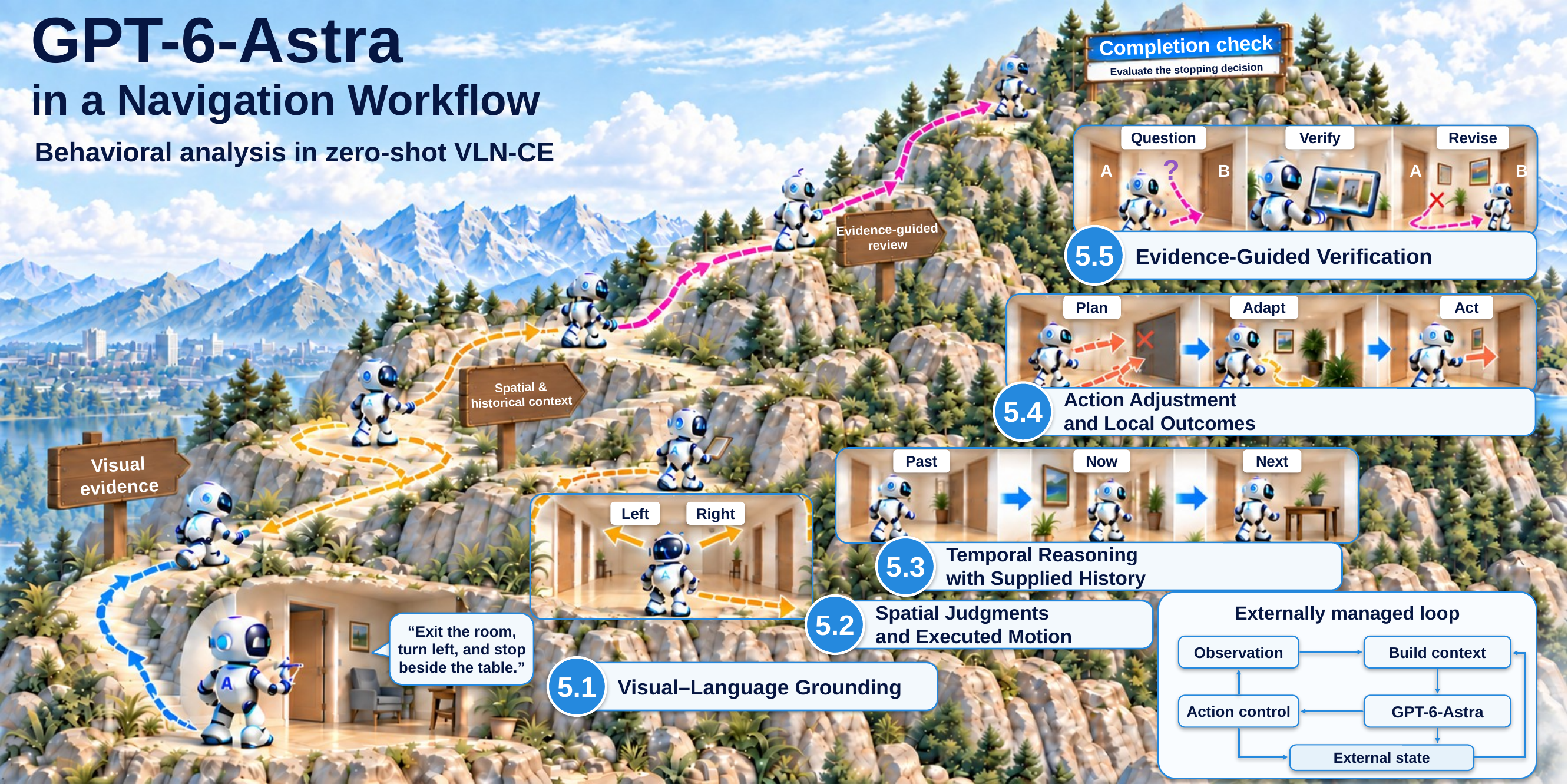}
    \captionsetup{type=figure}
    \caption{\textbf{From interpreting instructions to completing navigation.} Within the evaluated workflow, GPT-6-Astra can match landmarks and earlier actions to instructions, seek additional evidence, and revise uncertain judgments. These judgments do not always lead to progress or appropriate stopping. The five perspectives (Sections~5.1--5.5) organize this behavioral analysis; the mountain is a conceptual illustration.}
    \label{fig:teaser} 
  \end{center}
  \vspace{3pt}
} 
\makeatother    
     
\begin{document} 
\maketitle
    
\begin{abstract}
We study GPT-6-Astra in a zero-shot Vision-and-Language Navigation in Continuous Environments (VLN-CE) system, where it interprets instructions, assesses its surroundings, and proposes actions.
The system uses a common observation--decision--execution workflow with direct model API calls, without a packaged agent harness or navigation-specific fine-tuning. In this workflow, each request receives selected observations, execution feedback, and retained progress records. Evaluation covers the complete system, including context management and action control.
We evaluate the system on 50 of the 100 R2R-CE val-unseen episodes used by Open-Nav. It achieves a success rate of 52.0\%, an SPL of 48.9\%, and an nDTW of 70.8\%.
Our analysis highlights three findings. First, recorded responses link landmarks and earlier actions to instructions using observations and supplied history. Second, reviews include requests for additional views and revisions of uncertain judgments. Third, the results suggest a gap between task understanding and autonomous completion: an unfinished crossing is recognized while rotation continues.
At termination, 36.0\% of episodes succeed with a workflow-accepted STOP, while another 16.0\% meet the distance criterion at the step limit.
These results highlight a central challenge: translating correct local judgments into sustained progress and appropriate stopping.
\end{abstract}  
  
\section{Introduction}
\label{sec:introduction}
We study how GPT-6-Astra uses visual observations and navigation history to decide where to go, what to inspect, and when to stop. Its recorded responses contain detailed judgments about which landmark matches an instruction, what earlier actions accomplished, and what evidence is still missing. Such judgments are useful when several targets look plausible or the effect of a previous action is unclear. Their value for navigation, however, depends on whether they help the agent choose and complete the next action. This motivates following the model's judgments through an entire navigation task.

Vision-and-Language Navigation (VLN) provides a setting for this analysis: an agent must follow a natural-language instruction through a visual environment~\cite{r2r}. In continuous environments (VLN-CE), it must execute turns and translations to follow the route and then stop at the destination~\cite{vlnce}. Recognizing an entrance does not establish that the agent has crossed it, and approaching a goal does not establish that it will stop appropriately. We therefore ask: \textbf{\emph{can GPT-6-Astra use the supplied observations and action history to guide an agent to the instructed destination and determine when to stop?}}

Prior zero-shot navigation systems combine language-model decisions with visual observations, maps, history, and feedback~\cite{zhou2024navgpt_verified,chen2024mapgpt_verified,qiao2025opennav,shi2025smartway,shi2025fastsmartway}. Recent work further explores trajectory imagination, targeted perception, and tool-based verification~\cite{wang2025dreamnav,xue2026profocus,li2026agenticnav}. These approaches illustrate how observations and external mechanisms support navigation decisions. NavBench also distinguishes navigation comprehension from execution~\cite{qiao2025navbench}. Building on this distinction, we examine what GPT-6-Astra judges and what the navigation system subsequently executes within the same episode.

We use an independently implemented, common observation--decision--execution workflow and evaluate the subset listed in Table~\ref{tab:dataset}. GPT-6-Astra interprets observations, assesses progress, proposes actions, and reviews arrival evidence through direct model API calls, without a packaged agent harness or navigation-specific fine-tuning. External code selects the context for each request, retains prior observations and judgments, schedules reviews, and constrains execution. We therefore report the performance of this complete system and interpret the model's responses in light of the evidence and control mechanisms available at each step.

Our analysis links saved observations and model responses to executed actions, trajectories, and completion outcomes. The five perspectives in Figure~\ref{fig:teaser} organize this evidence and support three findings, each connecting a model judgment to its observed consequences.

\textbf{\textit{(1) GPT-6-Astra shows strengths in linking landmarks and earlier actions to instructions.}}
For example, it distinguishes the ``second on the left'' from neighboring openings and uses later views to confirm that an earlier turn matched the route instruction. The latter response preserves when the turn occurred while clarifying what it accomplished. Across the evaluated episodes, the timing and order analyses each find supported evidence for every applicable check in 31 episodes (Section~\ref{sec:temporal}). These judgments use supplied history. They show that later evidence can establish an instruction match for an action that has already occurred, so the time of an event and the time of its confirmation need to be distinguished.

\textbf{\textit{(2) GPT-6-Astra shows strengths in seeking visual information and revising uncertain judgments.}}
It proposes moving past a door leaf, after which the executed movement reveals a hidden passage. In another case, additional views expose shelving and a back wall, leading it to reject an uncertain hallway candidate. Our analysis finds that 36 of 66 adjustment processes help overcome a local difficulty, while 48 of 75 verification processes resolve an uncertain judgment with supporting evidence (Sections~\ref{sec:adaptive}--\ref{sec:reflective}). These are system processes, with reviews scheduled by the workflow. They demonstrate the value of obtaining evidence that answers a specific question, while leaving open whether resolving that question leads to further route progress.

\textbf{\textit{(3) GPT-6-Astra shows limitations in translating task understanding into autonomous completion within this workflow.}}
In a stalled doorway sequence, the model correctly distinguishes a completed turn from a pending crossing, yet the system continues rotating without crossing. A gap between physical arrival and accepted completion also appears in the aggregate results. The system achieves 52.0\% SR, 48.9\% SPL, and 70.8\% nDTW, but only 36.0\% of episodes succeed with a workflow-accepted STOP; another 16.0\% meet the distance criterion at the step limit. Thus, supported judgments and successful endpoints do not by themselves establish effective task completion. Assessing this system requires checking whether the required motion follows a judgment and whether arrival leads to an appropriate stopping decision.

\section{Task and Navigation Workflow}
\label{sec:formulation}

\begin{figure*}[t]
\centering
\includegraphics[width=\textwidth]{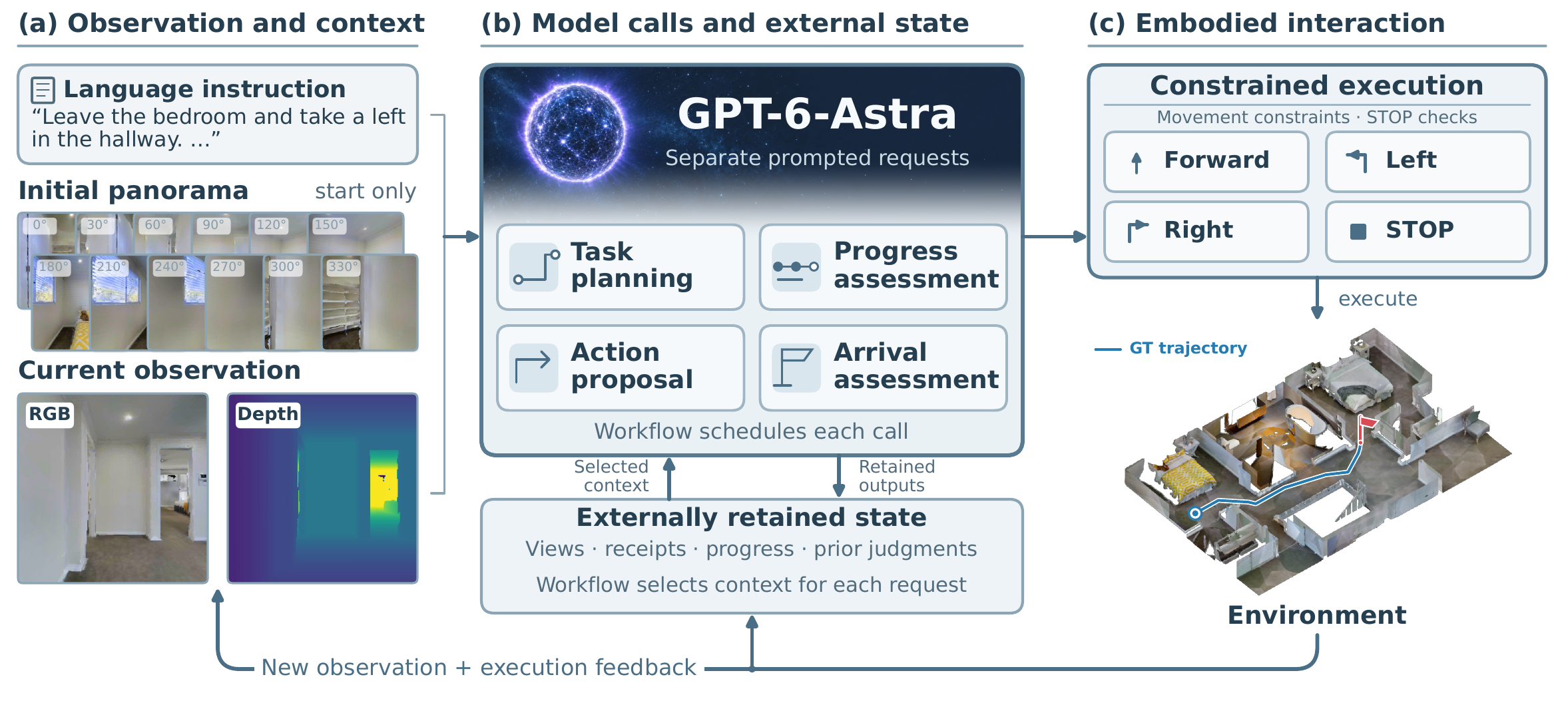}
\caption{\textbf{Model calls within an externally managed navigation loop.} (a) Observations supply scene evidence. (b) External state retains history and prior judgments; the workflow selects context and constructs separate calls for the four model roles. Outputs update retained records. (c) Action constraints and STOP checks precede execution, whose feedback updates the next request.}
\label{fig:framework}
\end{figure*}

\textbf{Task formulation.} Vision-and-Language Navigation in Continuous Environments (VLN-CE) requires an agent to follow a natural-language instruction $\mathcal{I}$ and reach the instructed destination through low-level actions~\cite{vlnce}. In the evaluated setting, the action space $\mathcal{A}$ comprises forward movement of 0.25\,m, left/right rotation of $15^\circ$, and STOP. The agent receives an initial panorama for orientation, followed by egocentric RGB/depth observations $\mathcal{O}_t$. Selected historical observations, executed actions, and execution feedback form the available history $\mathcal{H}_t$.

The navigation decision is
\begin{equation}
    a_t = \pi_{\mathrm{agent}}(\mathcal{I},\mathcal{O}_t,\mathcal{H}_t),
    \qquad a_t \in \mathcal{A},
    \label{eq:navigation}
\end{equation}
where $\pi_{\mathrm{agent}}$ denotes the complete navigation policy, combining GPT-6-Astra's proposals with external context management and action constraints. Executing $a_t$ produces the next observation and motion/collision feedback, which inform subsequent decisions. The GT route and endpoint are reserved for offline evaluation.

Following the instruction requires identifying the relevant landmarks, tracking completed actions, and obtaining more evidence when the next move is unclear. We examine whether these judgments lead to the required movement and an appropriate stopping decision.

\textbf{Division of responsibilities.} Figure~\ref{fig:framework} shows our implementation of the common observation--decision--execution loop using direct model API calls, without a packaged agent harness. The instruction and an initial scan of 12 headings establish the starting context; subsequent observations are frontal RGB-D. GPT-6-Astra serves four prompted roles: task planning, progress assessment, action proposal, and arrival assessment. Workflow code schedules these calls, selects their evidence, validates and retains structured outputs, and applies movement and completion constraints. The model can request additional saved images or propose a new viewpoint; the workflow supplies the selected images or executes an action after applying its constraints. The loop ends at a workflow-accepted STOP or the episode budget. Context selection and action control therefore contribute to the reported system performance alongside the model's judgments.

\section{Evaluation Setup}
\label{sec:setup}
\subsection{Dataset}
The evaluated subset comprises the first 50 episodes in ascending episode-ID order from the Open-Nav evaluation pool~\cite{qiao2025opennav}, which is also used by SmartWay and Fast-SmartWay~\cite{shi2025smartway,shi2025fastsmartway}. It spans nine reconstructed indoor scenes from the unseen-environment split. The evaluated episodes use the \texttt{R2R\_VLNCE\_v1-3} release, with corrected initial headings and matching GT trajectories~\cite{vlnceDataV13}. Table~\ref{tab:dataset} lists all evaluated episode IDs and their scene distribution.

\begin{table}[t]
\centering
\small
\setlength{\tabcolsep}{3pt}
\renewcommand{\arraystretch}{1.12}
\caption{\textbf{Composition of the evaluated subset.} All 50 episodes are grouped by scene; execution follows ascending episode-ID order across the subset.}
\label{tab:dataset}
\begin{tabularx}{\columnwidth}{@{}lrY@{}}
\toprule
Scene ID & \#Ep. & Episode IDs \\
\midrule
\texttt{2azQ1b91cZZ} & 9 & 11, 70, 116, 140, 166, 265, 330, 403, 513 \\
\texttt{EU6Fwq7SyZv} & 3 & 321, 348, 461 \\
\texttt{QUCTc6BB5sX} & 7 & 94, 150, 176, 190, 191, 423, 454 \\
\texttt{TbHJrupSAjP} & 8 & 171, 181, 259, 275, 308, 338, 377, 411 \\
\texttt{X7HyMhZNoso} & 4 & 218, 226, 244, 371 \\
\texttt{Z6MFQCViBuw} & 3 & 207, 432, 469 \\
\texttt{oLBMNvg9in8} & 2 & 232, 479 \\
\texttt{x8F5xyUWy9e} & 3 & 7, 187, 312 \\
\texttt{zsNo4HB9uLZ} & 11 & 13, 40, 42, 52, 156, 247, 362, 387, 439, 447, 516 \\
\midrule
\textbf{9 scenes} & \textbf{50} & \textbf{50 of 100 parent episodes} \\
\bottomrule
\end{tabularx}
\end{table}

\subsection{Context Management and Action Interface}
Each generation is a separate chat request containing a role-specific system prompt and one newly constructed multimodal user message. The implementation does not replay the full assistant-message history or pass reasoning-state continuation items between requests. Earlier observations, execution receipts, task progress, and selected model judgments are retained externally and supplied in later requests. Decisions can therefore use navigation history, with the workflow determining which parts of that history remain available.

Ordinary action calls receive up to four selected observation frames and an eight-action receipt window, with the task-activation view retained when needed. Progress reviews can draw on the episode archive, event records, prior questions, and validation feedback; arrival reviews additionally receive the two most recent review records. The model can request saved images from a supplied catalog, subject to workflow selection limits. Structured event updates are validated and persisted by code, including prerequisite-order checks. These supplied records and constraints support the temporal judgments analyzed in Section~\ref{sec:temporal}.

The action controller applies depth and failed-heading restrictions, can select alignment or recovery actions, and checks completion before accepting STOP. We distinguish model proposals from executed actions using the saved receipts. Offline GT trajectories, endpoints, and benchmark distance-to-goal are excluded from online decisions.

\subsection{Model and Execution Settings}
The saved request identifies the model as \texttt{gpt-6-astra}, accessed through the OpenAI-compatible chat interface with \texttt{reasoning\_effort=medium}. The recorded output-token limit is 8,192 per generation. Initial observation provides 12 panoramic headings. Online actions are forward movement of 0.25\,m, left/right rotation of $15^\circ$, and STOP, under a 100-step episode budget. The configuration records seed 0; this does not establish deterministic API behavior. We analyze one execution of this configuration, without repeated trials, model substitutions, or component ablations. Here, ``zero-shot'' means no navigation-specific fine-tuning of GPT-6-Astra; the subset was not held out from all workflow development.

\subsection{Metrics and Evidence}
We report terminal navigation error (NE), success rate (SR), oracle success rate (OSR), success weighted by path length (SPL), and normalized dynamic time warping (nDTW)~\cite{vlnce,ilharco2019ndtw_verified}. These are system-level outcomes. The success radius is 3\,m; OSR tests whether the trajectory ever reaches that neighborhood, whereas SR concerns its endpoint. SPL and nDTW summarize path efficiency and reference-path agreement, respectively. The saved evaluator also counts successful endpoints reached at the budget limit. We use \emph{autonomous completion} to denote a successful endpoint accompanied by a workflow-accepted STOP; this remains a system-level outcome. Evidence comprises final metrics, original images, model outputs, executed-action receipts, and offline GT. Action index $t$ denotes the action from observation frame $t$ to $t+1$. Section~\ref{sec:analysis} combines analyses of all evaluated episodes with cases that illustrate the observed patterns. Its labels indicate what the saved evidence supports; they do not measure isolated model accuracy.

\section{Quantitative Results}
\label{sec:results}
\subsection{Overall Navigation Performance}
\label{sec:sota}
Across all 50 evaluated episodes, the GPT-6-Astra-based system achieves an NE of 4.34\,m, nDTW of 70.79\%, OSR of 62.00\%, SR of 52.00\%, and SPL of 48.90\% (Table~\ref{tab:sota}). These results summarize one recorded execution on the fixed subset in Table~\ref{tab:dataset}: 26 episodes end within the success neighborhood. They characterize the complete observation, context-management, decision, and execution workflow described in Section~\ref{sec:setup}.

Table~\ref{tab:sota} organizes published results along three independent attributes: learning regime, evaluation cohort, and observation setting. Full denotes all 1,839 R2R-CE val-unseen episodes; ON-100 denotes the 100-episode pool used by Open-Nav and subsequent work~\cite{qiao2025opennav,shi2025smartway}. Other sampled or unspecified cohorts are identified separately. Pano. and Mono. distinguish panoramic and monocular observations, while F3+P denotes three forward-facing views with on-demand panoramas, as used by Fast-SmartWay and C$^{2}$Nav~\cite{shi2025fastsmartway,zheng2026c2nav}. The evaluated system uses frontal RGB-D observations after an initial panorama and is listed under Mono. The evaluated subset uses the corrected \texttt{R2R\_VLNCE\_v1-3} release. Differences in episode coverage, data versions, observation access, and action budgets make the table a contextual comparison rather than a matched ranking.

\begin{table*}[tp]
\centering
\small
\setlength{\tabcolsep}{1.5pt}
\renewcommand{\arraystretch}{1.00}
\caption{\textbf{R2R-CE val-unseen navigation results.} Val unseen denotes the cohort; View denotes the observation setting. Our GPT-6-Astra-based system uses 50 episodes from the Open-Nav 100-episode pool~\cite{qiao2025opennav}. NE is in meters; other metrics are percentages. Cohorts and workflows differ, so these results are not a matched model ranking.}
\label{tab:sota} 
\begin{tabular*}{\textwidth}{@{\extracolsep{\fill}}llccrrrrr@{}}
\toprule
Method & Source & \shortstack{Val unseen} & \shortstack{View} & NE $\downarrow$ & nDTW $\uparrow$ & OSR $\uparrow$ & SR $\uparrow$ & SPL $\uparrow$ \\
\midrule 
\rowcolor{black!6}
\multicolumn{9}{l}{\emph{VLN-CE supervised learning}} \\
CMA~\cite{dcvln} & CVPR'2022 & ON-100 & Pano. & 6.92 & 50.77 & 45.00 & 37.00 & 32.17 \\
RecBERT~\cite{dcvln} & CVPR'2022 & ON-100 & Pano. & 5.80 & 54.81 & 57.00 & 48.00 & 43.22 \\
BEVBert~\cite{an2023bevbert} & ICCV'2023 & ON-100 & Pano. & 5.13 & 61.40 & 64.00 & 60.00 & 53.41 \\
ETPNav~\cite{an2025etpnav_verified} & TPAMI'2025 & ON-100 & Pano. & 5.16 & 61.16 & 63.00 & 58.00 & 52.19 \\
\midrule
ScaleVLN~\cite{wang2023scaling} & ICCV'2023 & Full & Pano. & 4.80 & -- & -- & 55.00 & 51.00 \\
ETPNav~\cite{an2025etpnav_verified} & TPAMI'2025 & Full & Pano. & 4.71 & -- & 65.00 & 57.00 & 49.00 \\
BEVBert~\cite{an2023bevbert} & ICCV'2023 & Full & Pano. & 4.57 & -- & 67.00 & 59.00 & 50.00 \\
HNR~\cite{wang2024hnr_verified} & CVPR'2024 & Full & Pano. & 4.42 & -- & 67.00 & 61.00 & 51.00 \\
Energy~\cite{liu2024energy} & NeurIPS'2024 & Full & Pano. & 4.69 & -- & 65.00 & 58.00 & 50.00 \\
g3D-LF~\cite{wang2025g3dlf_verified} & CVPR'2025 & Full & Pano. & 4.53 & -- & 68.00 & 61.00 & 52.00 \\
\midrule
NaVid~\cite{zhang2024navid_rss} & RSS'2024 & Full & Mono. & 5.47 & -- & 49.10 & 37.40 & 35.90 \\
Uni-NaVid~\cite{zhang2025uninavid_rss} & RSS'2025 & Full & Mono. & 5.58 & -- & 53.30 & 47.00 & 42.70 \\
NaVILA~\cite{cheng2025navila_rss} & RSS'2025 & Full & Mono. & 5.22 & -- & 62.50 & 54.00 & 49.00 \\
Aux-Think~\cite{wang2025auxthink_neurips} & NeurIPS'2025 & Full & Mono. & 5.88 & -- & 54.90 & 49.70 & 41.70 \\
Dynam3D~\cite{wang2025dynam3d_neurips} & NeurIPS'2025 & Full & Mono. & 5.34 & -- & 62.10 & 52.90 & 45.70 \\
StreamVLN~\cite{wei2026streamvln_icra} & ICRA'2026 & Full & Mono. & 4.98 & -- & 64.20 & 56.90 & 51.90 \\
D3D-VLP~\cite{wang2026d3dvlp_verified} & CVPR'2026 & Full & Mono. & 4.73 & -- & 67.20 & 61.30 & 56.10 \\
\midrule
\rowcolor{black!6}
\multicolumn{9}{l}{\emph{Zero-shot VLN-CE}} \\
HSGM~\cite{li2026hsgm_verified} & CVPR'2026 & Full & Pano. & 5.42 & -- & 58.70 & 47.90 & 32.80 \\
LXMERT~\cite{dcvln} & CVPR'2022 & ON-100 & Pano. & 10.48 & 18.73 & 22.00 & 2.00 & 1.87 \\
MapGPT-CE-GPT4o$^{\dagger}$~\cite{chen2024mapgpt_verified,shi2025smartway} & ACL'2024 & ON-100 & Pano. & 8.16 & -- & 21.00 & 7.00 & 5.04 \\
DiscussNav-GPT4$^{\dagger}$~\cite{long2024discussnav_verified,shi2025smartway} & ICRA'2024 & ON-100 & Pano. & 7.77 & 42.87 & 15.00 & 11.00 & 10.51 \\
NavGPT-CE-GPT4$^{\dagger}$~\cite{zhou2024navgpt_verified,dai2026evonav} & AAAI'2024 & ON-100 & Pano. & 8.37 & -- & 26.90 & 16.30 & 10.20 \\
Open-Nav-GPT4~\cite{qiao2025opennav} & ICRA'2025 & ON-100 & Pano. & 6.70 & 45.79 & 23.00 & 19.00 & 16.10 \\
Open-Nav-Gemini-2.5-pro~\cite{qiao2025opennav,dai2026evonav} & ICRA'2025 & ON-100 & Pano. & 7.28 & 49.51 & 30.00 & 23.00 & 19.90 \\
Open-Nav-Gemini-3-Flash$^{r}$~\cite{qiao2025opennav} & ICRA'2025 & ON-100 & Pano. & 6.42 & 54.82 & 42.00 & 32.00 & 27.65 \\
Three-Step Nav-GPT-5~\cite{zheng2026threestepnav} & AISTATS'2026 & ON-100 & Pano. & 5.87 & 57.70 & 39.00 & 34.00 & 29.12 \\
LaViRA-GPT-4o~\cite{ding2026lavira_icra} & ICRA'2026 & ON-100 & Pano. & 6.43 & -- & 43.30 & 36.00 & 28.30 \\
LaViRA-Gemini-2.5-pro~\cite{ding2026lavira_icra} & ICRA'2026 & ON-100 & Pano. & 6.54 & -- & 48.70 & 38.30 & 28.30 \\
EvoNav-Gemini-2.5-pro~\cite{dai2026evonav} & CVPR'2026 & ON-100 & Pano. & 5.04 & 62.38 & 51.00 & 43.00 & 37.77 \\
SmartWay-GPT-5.5$^{r}$~\cite{shi2025smartway} & IROS'2025 & ON-100 & Pano. & 5.16 & 58.64 & 60.00 & 44.00 & 35.04 \\
AgenticNav-Gemini-2.5-pro~\cite{li2026agenticnav} & arXiv'2026 & ON-100 & Pano. & 5.91 & 48.73 & 63.00 & 49.00 & 33.20 \\
AgenticNav-GPT-5.5~\cite{li2026agenticnav} & arXiv'2026 & ON-100 & Pano. & 5.19 & 63.41 & 65.00 & 55.00 & 48.41 \\
SpatialAnt$^{p}$~\cite{zhang2026spatialant_verified} & arXiv'2026 & S-100 & Pano. & 4.42 & 69.50 & 76.00 & 66.00 & 54.40 \\
HarnessVLN-GPT-5.5~\cite{chen2026harnessvln_verified} & arXiv'2026 & - & Pano. & 4.01 & -- & 72.70 & 60.80 & 43.50 \\
\midrule
Fast-SmartWay-GPT-4o~\cite{shi2025fastsmartway} & arXiv'2025 & ON-100 & F3+P & 7.72 & 51.83 & -- & 27.75 & 24.95 \\
C$^{2}$Nav-GPT-5.5~\cite{zheng2026c2nav} & arXiv'2026 & ON-100 & F3+P & 7.20 & -- & 54.00 & 44.00 & 29.00 \\
\midrule
CA-Nav~\cite{chen2025canav_verified} & TPAMI'2025 & Full & Mono. & 7.58 & -- & 48.00 & 25.30 & 10.80 \\
AO-Planner~\cite{chen2025aoplanner_verified} & AAAI'2025 & Full & Mono. & 6.95 & -- & 38.30 & 25.50 & 16.60 \\
DreamNav~\cite{wang2025dreamnav} & arXiv'2025 & 613 traj. & Mono. & 7.06 & -- & 41.00 & 32.80 & 29.00 \\
GC-VLN~\cite{yin2025gcvln_corl} & CoRL'2025 & Full & Mono. & 7.30 & -- & 41.80 & 33.60 & 16.30 \\
\midrule 
\rowcolor{black!6} 
\textbf{GPT-6-Astra + our workflow} & Report'2026 & 50 of ON-100 & Mono. & \textbf{4.34} & \textbf{70.79} & \textbf{62.00} & \textbf{52.00} & \textbf{48.90} \\
\bottomrule
\end{tabular*}
\vspace{3pt}
\begin{minipage}{\textwidth}\footnotesize
\textit{Evaluation sets.} Full: all 1,839 val-unseen episodes. ON-100: the reported Open-Nav 100-episode protocol. S-100: 100 sampled episodes with identity to ON-100 unverified. 50 of ON-100: the evaluated subset (Table~\ref{tab:dataset}). A dash denotes an unspecified episode cohort. DreamNav reports 613 trajectories; instruction multiplicity is unspecified. LaViRA scores are means over three runs; both variants use Qwen2.5-VL-32B for visual grounding. Fast-SmartWay scores are means over four runs. Missing metrics remain ``--''; bold identifies this report's system result.\par
\textit{Inputs and settings.} Pano.: panoramic observations; Mono.: a monocular egocentric camera; F3+P: three forward-facing views plus on-demand panoramas. Fast-SmartWay also uses an initial panorama. $^{p}$SpatialAnt reconstructs scenes and uses simulator depth for R2R-CE waypoint prediction. $^{\dagger}$Extra annotations/oracle signals in the source comparison. $^{r}$Reproduced by AgenticNav's authors.
\end{minipage} 
\end{table*}

Among zero-shot ON-100 results, Three-Step Nav reports 34.00\% SR and 29.12\% SPL with panoramic observations~\cite{zheng2026threestepnav}. Fast-SmartWay and C$^{2}$Nav report 27.75\%/24.95\% and 44.00\%/29.00\% SR/SPL, respectively, with F3+P observations; the former averages four runs. AgenticNav with GPT-5.5 reports 55.00\% SR and 48.41\% SPL~\cite{li2026agenticnav}. The evaluated system's SPL is numerically close to the latter result, while its SR is lower. The different cohorts and workflows preclude attributing these differences to the language model or observation interface alone.

Supervised results provide a further reference under their own evaluation conditions. On the full split, panoramic g3D-LF reports 61.00\% SR and 52.00\% SPL, while monocular D3D-VLP reports 61.30\% and 56.10\%~\cite{wang2025g3dlf_verified,wang2026d3dvlp_verified}. These papers also supply the full-split baseline blocks in Table~\ref{tab:sota}; sampled supervised results remain separate. The numerical proximity of selected scores is insufficient to establish that this system matches a trained navigator across the full benchmark.

Within the evaluated subset, 31 trajectories enter the success neighborhood at least once, but five end outside it. This 10-percentage-point OSR--SR gap shows that reaching the goal region does not ensure a successful endpoint. We next examine whether successful endpoints coincide with an accepted stopping decision.

\subsection{Arrival and Stopping Outcomes}
\label{sec:stop}
Successful endpoints do not always coincide with workflow-accepted stopping (Table~\ref{tab:stopping}). The workflow accepts STOP in 21 episodes: 18 succeed and three fail the distance criterion. The remaining 29 episodes terminate at the action budget, including eight successful endpoints and 21 failures. Successful navigation with a workflow-accepted STOP thus occurs in $18/50=36.0\%$ of the evaluated episodes; budget-terminated successes contribute another 16.0 percentage points to the reported 52.0\% SR. Accepted stopping combines model-based arrival assessment with workflow checks, so this measure is not the model's standalone completion accuracy.

\begin{table}[t]
\centering
\small
\caption{\textbf{Endpoint success and workflow-accepted stopping.} Counts cover all 50 episodes. Success uses the endpoint distance criterion; accepted STOP combines model assessment with workflow checks. Budget termination can also yield a successful endpoint.}
\label{tab:stopping}
\begin{tabular}{lrrr}
\toprule
Termination & Success & Failure & Total \\
\midrule
Workflow-accepted STOP & 18 & 3 & 21 \\
Budget termination & 8 & 21 & 29 \\
\midrule
Total & 26 & 24 & 50 \\
\bottomrule
\end{tabular}
\end{table}

The unsuccessful accepted STOPs occur in EP7, EP371, and EP377, at NE values of 3.04, 3.37, and 4.07\,m. Their distances show varying degrees of endpoint error, including one close to the 3\,m threshold. Conversely, EP423 ends only 0.68\,m from the goal at budget exhaustion while its last online decision still requests forward movement. These outcomes illustrate the completion gap in the third finding: stopping outside the success region or reaching it without an accepted completion decision. The route and observation evidence in Section~\ref{sec:spatial} further examines the latter case.

\subsection{Efficiency and Inference Workload}
\label{sec:cost}
Path efficiency and execution workload capture different aspects of navigation. The system records 4,040 executed steps, averaging 80.8 steps and 8.97\,m of translational path per episode. These comprise 1,808 forward actions, 2,182 rotations, and 50 terminal STOPs, including budget-generated STOPs. Rotations account for 54.0\% of the action count. They can support alignment and evidence acquisition while adding no translational path length, so the reported SPL of 48.90\% does not imply low action or inference cost.

\begin{figure*}[t]
\centering
\includegraphics[width=\textwidth]{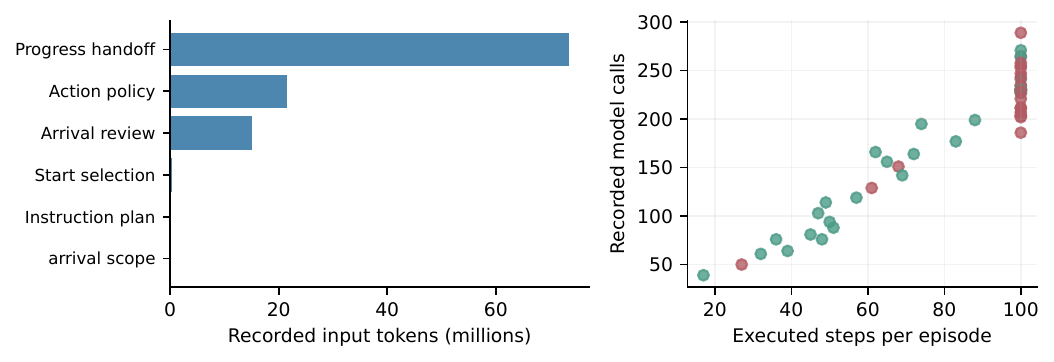}
\caption{\textbf{Inference workload across all 50 episodes.} Left: recorded input tokens by call role. Right: model generations versus executed steps; green denotes endpoint success and red denotes failure. Multiple reviews can precede one action, and tokens include repeated context. Counts describe recorded workload rather than billed cost.}
\label{fig:workload}
\end{figure*}

The traces contain 9,208 recorded model generations, including recorded format-rejected outputs. This corresponds to 2.28 generations per executed step on average, with a median of 205.5 generations per episode and a range of 39--289. Figure~\ref{fig:workload} relates these counts to executed steps and separates input-token usage by call role. Progress handoff and arrival review together consume 80.2\% of recorded input tokens, showing that maintaining progress and assessing completion dominate this workflow's input workload.

Available usage fields total 110.41 million input tokens and 3.24 million output tokens; reasoning-token metadata are not added again to the completion total. Recorded generation latencies sum to 22.55 hours, a cumulative call-time measure rather than end-to-end runtime. The logs do not establish an attributable bill or complete cache-discount history. These measurements quantify the workload of the evaluated system without isolating which reviews were necessary. Section~\ref{sec:analysis} examines what the model establishes from the available evidence and whether subsequent actions advance the task.

\section{Behavioral and Failure Analysis}
\label{sec:analysis}
Five overlapping perspectives examine the strengths and limitations behind the three findings (Figure~\ref{fig:teaser}). Sections~\ref{sec:grounding}--\ref{sec:temporal} relate target and event judgments to observations, supplied history, and executed motion. Sections~\ref{sec:adaptive}--\ref{sec:reflective} examine information seeking and judgment revision. Across these analyses, we follow supported judgments through subsequent movement and completion outcomes. Table~\ref{tab:capabilities} summarizes observed strengths, unresolved difficulties, and evidence boundaries. Model responses show expressed judgments, external records show retained context, and execution receipts show what physically happened. These perspectives do not define independent capability scores.

\begin{table*}[t]
\centering
\small
\setlength{\tabcolsep}{5pt}
\renewcommand{\arraystretch}{1.15}
\caption{\textbf{Behavioral evidence within the evaluated workflow.} Five overlapping perspectives connect contextual model judgments with observed execution. Cases and process audits describe this system; they do not isolate intrinsic model capabilities.}
\label{tab:capabilities}
\begin{tabularx}{\textwidth}{p{.19\textwidth}YY}
\toprule
Perspective & Observed strengths & Limitations and evidence boundaries \\
\midrule
Visual--language grounding & EP218 distinguishes a second-left doorway from a separate forward opening; EP423 identifies the bathtub. & EP469 observes relevant-looking display panes while the route-qualified destination remains unresolved. \\
Spatial judgments and motion & EP423 follows the two-hallway route; EP218's translated viewpoint exposes an occluded doorway. & EP116 does not complete a visible doorway crossing; executed movement is also constrained by the controller. \\
Temporal reasoning with history & Given retained records and observations, EP423 relates later confirmation to an earlier turn; EP116 distinguishes turning from crossing. & These judgments use supplied history and order checks; they do not establish unaided memory. \\
Action adjustment & Across the cohort, 36 of 66 audited problem-handling processes show local relief, covering 24 episodes. & 29 processes remain unresolved; local relief does not isolate a model contribution or guarantee navigation success. \\
Evidence-guided verification & 48 of 75 audited verification processes clarify an uncertain judgment with supporting evidence, covering 33 episodes. & 22 processes retain uncertainty and five are unassessable; the workflow initiates reviews. \\
\bottomrule
\end{tabularx}
\end{table*}

\subsection{Visual--Language Grounding}
\label{sec:grounding}
\textbf{Identifying the destination requires more than recognizing its category.}
We examine whether the saved evidence supports both the stated object or scene category and the qualifiers that identify the intended destination. The analysis covers all 50 completed episodes, with one \emph{terminal destination reference} per original instruction. Repeated observations of the same target count once; intermediate landmarks are outside this inventory. The preliminary offline, assistant-assisted audit combines saved target judgments, execution receipts, GT/executed paths, and 319 selected original RGB frames across the nine scenes. The judgments are conditioned on workflow-selected views and context. Model confidence and internal ``supported'' flags are claims to inspect, not annotation labels; GT is used only offline.

We distinguish \textbf{\textit{(1) category recognition}}, whether a visually identified candidate has the stated object or scene category; \textbf{\textit{(2) relational grounding}}, whether observations support its instruction-qualified spatial or route relation; and \textbf{\textit{(3) target-instance correspondence}}, whether the selected physical reference is consistent with both the visual context and the GT destination route. Each dimension is labeled supported (S), reference-inconsistent (C), or unassessable (U). U includes targets not established in the recorded views, unresolved alternatives, and insufficient independent evidence. A target not reached or not identified is therefore not automatically a recognition error. Composite references require their relevant components: seeing a bathroom, for example, does not establish the separate toilet room named in the instruction.

\begin{table}[t]
\centering
\small
\setlength{\tabcolsep}{4pt}
\renewcommand{\arraystretch}{1.1}
\caption{\textbf{Terminal-reference evidence across all 50 episodes.} Preliminary offline assistant-assisted audit, one reference per episode. S: supported; C: reference-inconsistent; U: unassessable. The last column uses all 50 episodes, not only assessable cases.}
\label{tab:grounding_census}
\begin{tabularx}{\columnwidth}{@{}Yrrrr@{}}
\toprule
Dimension & S & C & U & S / 50 \\
\midrule
\textbf{\textit{Category recognition}} & 33 & 0 & 17 & 66.0\% \\
\textbf{\textit{Relational grounding}} & 23 & 0 & 27 & 46.0\% \\
\textbf{\textit{Target-instance correspondence}} & 18 & 1 & 31 & 36.0\% \\
\bottomrule
\end{tabularx}
\vspace{2pt}
\begin{minipage}{\columnwidth}\footnotesize
Assessable counts (S+C) are 33, 23, and 19. U includes targets not established in the saved views and unresolved reference identity. C denotes a candidate/GT-route inconsistency, not an isolated perceptual cause. These are evidence-coverage counts, not validated accuracy scores.
\end{minipage}
\end{table}

\textbf{The intended target is established less often than its category.}
Table~\ref{tab:grounding_census} reports support in 33/50 episodes for category recognition, 23/50 for relations, and 18/50 for target-instance correspondence. Fifteen episodes have category support without independently supported instance correspondence. This gap includes ambiguity and limited visibility as well as an observed candidate/reference-route mismatch; it is not a count of 15 semantic errors. The 31 unassessable instance judgments remain in the denominator. These percentages measure evidence coverage within this cohort, not validated capability accuracy. Different visibility and assessability also prevent interpreting their difference as a calibrated ranking of the three abilities. No independent human annotation or inter-rater study has been performed.

\begin{figure*}[tp]
\centering
\includegraphics[width=\textwidth]{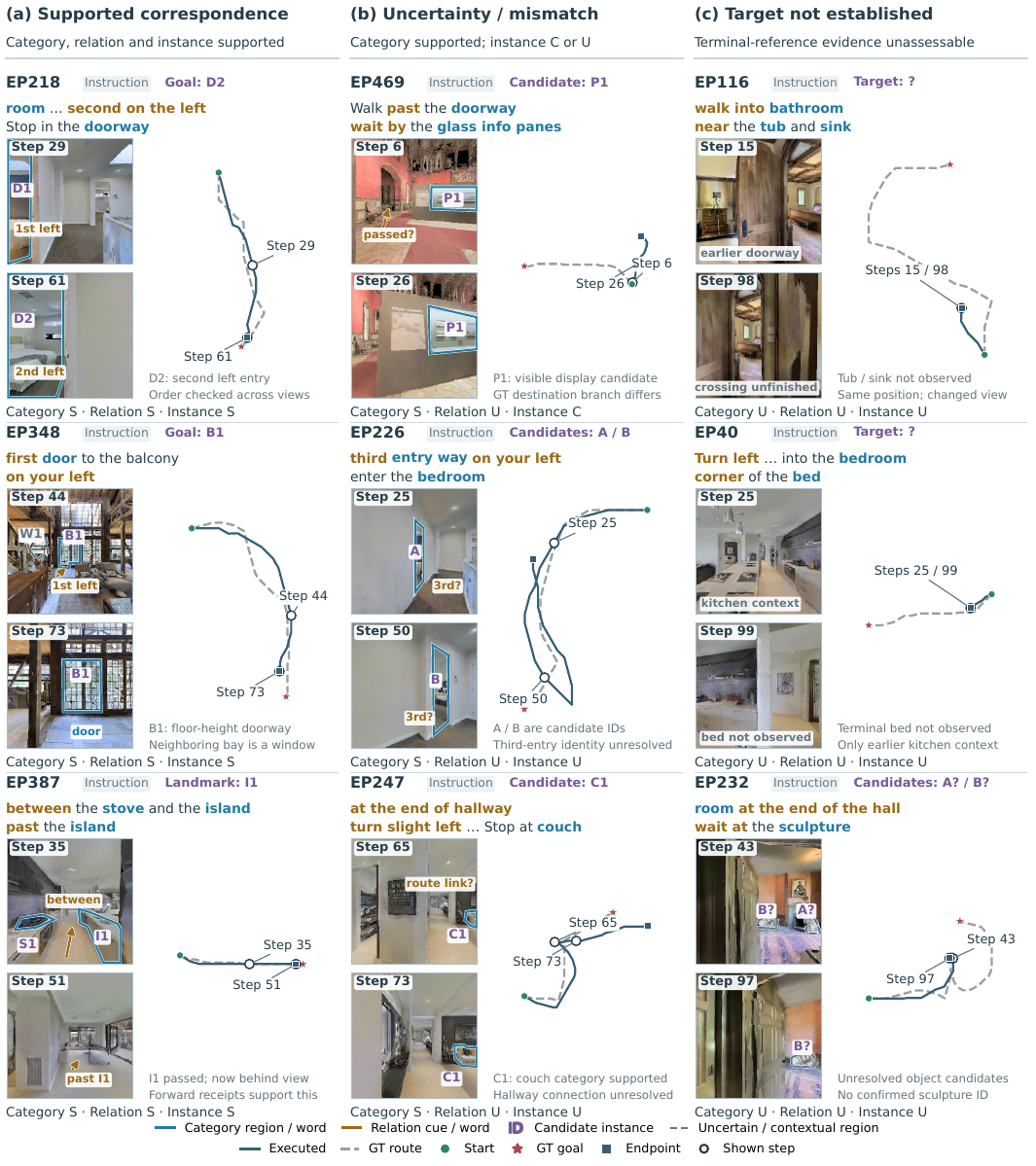}
\caption{\textbf{Visual--language grounding in navigation.} Panels show supported correspondence (a), uncertain or inconsistent reference (b), and targets not established (c). \textcolor[HTML]{217BA5}{Blue}, \textcolor[HTML]{9B6A16}{ochre}, and \textcolor[HTML]{755A99}{purple} link category words/regions, relational cues, and candidate IDs, respectively. Step $t$ denotes the observation after $t$ actions. S/C/U follow Table~\ref{tab:grounding_census}.}
\label{fig:grounding_census_cases}
\end{figure*}

\textbf{\textit{Qualitative grounding evidence.}}
Figure~\ref{fig:grounding_census_cases} links instruction phrases, visual evidence, and trajectories across three settings. In panel (a), EP218 distinguishes the ``second on the left'' from neighboring openings; EP348 separates a balcony door from adjacent windows, and EP387 establishes passage between the stove and island and beyond the island. Visual and route evidence jointly support all three grounding dimensions. In panel (b), EP469 recognizes the glass information panes, but the selected display lies on a different branch from the GT destination route while the doorway qualifier remains unresolved. This is a reference-route inconsistency, with its cause undetermined. EP226 and EP247 likewise recognize the relevant category but do not resolve the required entry order or hallway-turn relation; their instance labels are U, whereas EP469 alone is C. In panel (c), EP116 stalls before observing the terminal tub and sink, EP40 remains in the kitchen-side area without observing the bedroom bed, and EP232 leaves the sculpture unresolved among candidate objects. All three dimensions are U for these cases. Supported cases illustrate the first finding; unresolved cases show why category recognition alone is insufficient to identify the instructed target. Missing evidence remains distinct from recognition error.

\subsection{Spatial Judgments and Executed Motion}
\label{sec:spatial}
Seeing a passage, crossing it, and following the instructed route are separate steps. Three perspectives organize the cases: \textbf{\textit{(1) position and heading}}, relative to an entrance or travel lane; \textbf{\textit{(2) local passage}}, relating visible openings to depth, obstacles, and movement constraints; and \textbf{\textit{(3) spatial connectivity}}, distinguishing an onward corridor from a side room or another branch. These overlap with the instruction-qualified relations in Section~\ref{sec:grounding}. We examine model judgments, controller-selected or constrained actions, and recorded motion together to characterize progress and stalled crossings; they do not isolate spatial-perception accuracy.

\begin{figure*}[tp]
\centering
\includegraphics[width=\textwidth]{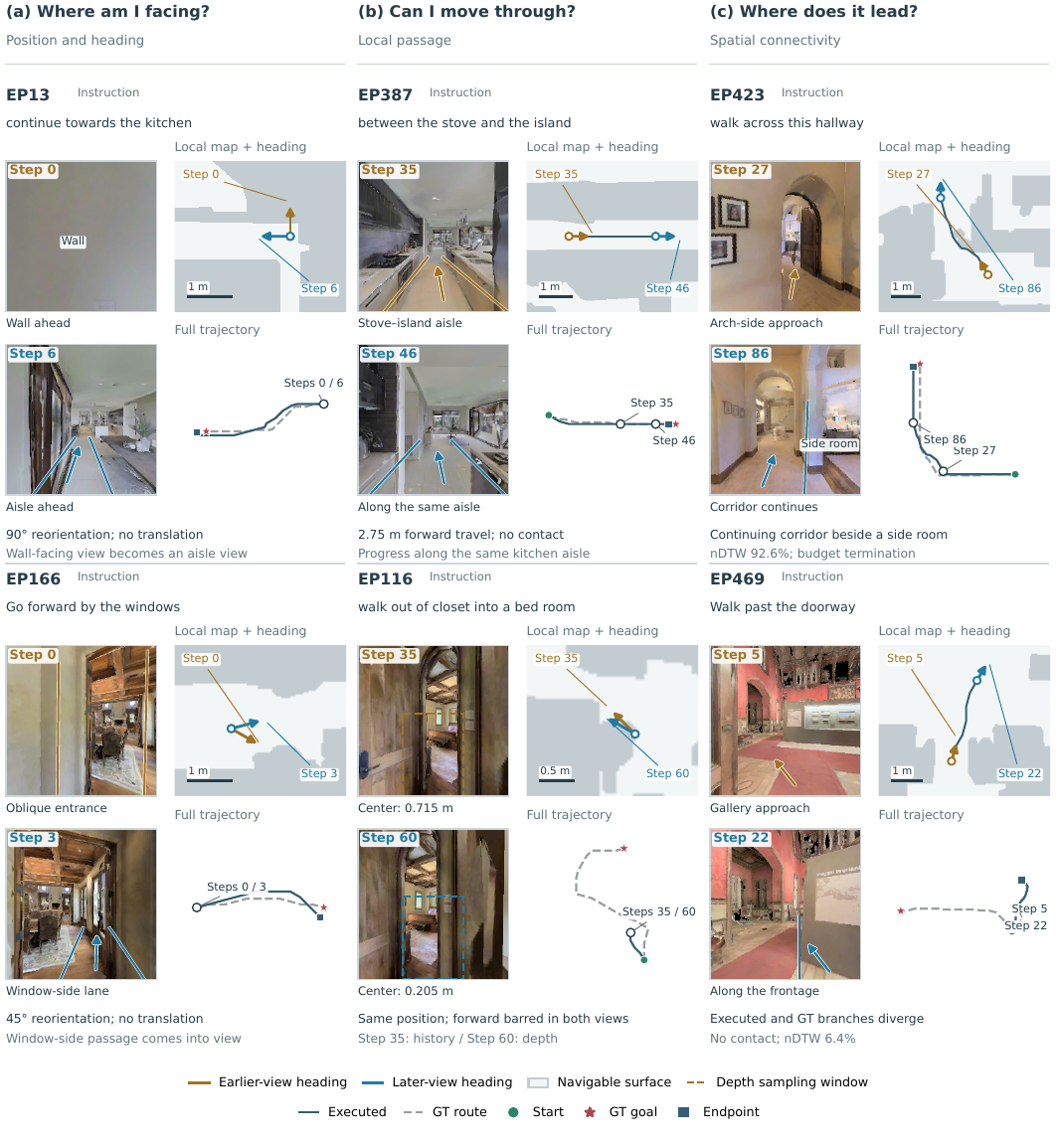}
\caption{\textbf{Spatial judgments and executed motion.} Columns show position and heading (a), local passage (b), and spatial connectivity (c), including controller-mediated movement. Paired RGB views match the labeled poses. Image arrows highlight passage cues; map arrows show heading. Dashed boxes mark the depth sampling window; values are its recorded 10th-percentile depth. Navmesh crops and GT routes are offline references. Local maps include scale bars; full trajectories use independent scales.}
\label{fig:grounding}
\end{figure*}

\textbf{\textit{From facing a passage to following its connection.}}
Figure~\ref{fig:grounding} makes these distinctions visible. In panel (a), EP13 turns 90$^{\circ}$ from a wall toward the kitchen aisle, and EP166 turns 45$^{\circ}$ toward the window-side passage; both pairs share a position, isolating the change in heading. EP166's initial alignment is executed by the panorama-bootstrap controller, so the examples characterize the complete workflow. In panel (b), EP387 advances 2.75\,m between the stove and island without contact. EP116 instead remains at the same doorway position: Step~35 has 0.715\,m center depth but a failed-heading restriction, whereas Step~60 has only 0.205\,m center depth and a depth restriction. The visible bedroom does not establish a completed crossing, and these records do not show that bypassing either restriction would be safe. In panel (c), EP423's views expose a continuing corridor beside a separate room, consistent with its reference-route alignment; its episode nevertheless ends at the budget limit. EP469 moves without contact along another gallery branch, leaving the instructed route unresolved. Together, the cases connect viewpoint geometry, admissible motion, and spatial connectivity without treating a recognizable destination or collision-free movement as sufficient route understanding.

\subsection{Temporal Reasoning with Supplied History}
\label{sec:temporal}
GPT-6-Astra's recorded reviews include distinctions between completed and unfinished events under the supplied history. We examine these judgments through \textbf{\textit{(1) event timing}}, separating physical occurrence from later confirmation; \textbf{\textit{(2) order constraints}}, distinguishing prerequisites such as entering a room before the next turn; and \textbf{\textit{(3) event continuity}}, explicitly referring to a past occurrence while distinguishing a later visit or a different instruction match. The workflow retains event records, selects historical views and receipts, and applies prerequisite-order checks. The analysis therefore concerns judgments made with this support, not autonomous retention of the navigation history.

\textbf{\textit{Evidence across all evaluated episodes.}}
We conduct a preliminary offline, assistant-assisted audit of all 50 episodes. The original instructions yield 172 motion clauses and 122 adjacent clause pairs, including unreached clauses; terminal waiting and stopping are treated separately. Saved model responses are checked against original views and execution receipts. Timing checks whether a stated occurrence and interval are compatible with the recorded motion, without validating an exact earliest boundary or the intended route. Continuity uses 132 clauses whose first occurrence claim has a progress-review opportunity at least ten actions later. It requires an explicit reference to the identifiable past event; an unchanged controller record alone is insufficient. Repeated answers about one unit count once, and the model's own support flags are not correctness labels.

\begin{table}[t]
\centering
\caption{\textbf{Temporal judgments under supplied context.} The audit covers all 50 episodes. $N$ counts motion clauses, adjacent order pairs, or eligible continuity checks. S: supported; C: contradicted; U: unassessable.}
\label{tab:temporal_census}
\small
\setlength{\tabcolsep}{3.7pt}
\begin{tabular}{@{}lrrrrr@{}}
\toprule
Dimension & $N$ & S & C & U & S/$N$ \\
\midrule
\textbf{\textit{Event timing}} & 172 & 146 & 0 & 26 & 84.9\% \\
\textbf{\textit{Order constraints}} & 122 & 100 & 0 & 22 & 82.0\% \\
\textbf{\textit{Event continuity}} & 132 & 61 & 0 & 71 & 46.2\% \\
\bottomrule
\end{tabular}
\vspace{2pt}
\parbox{\columnwidth}{\scriptsize 
Timing covers 50 episodes; order covers 44 with multiple motion clauses; continuity covers 47 with a later review at least 10 actions after the first occurrence claim. Repeated answers count once per unit. S/$N$ measures evidence coverage, not accuracy.}
\end{table}

Table~\ref{tab:temporal_census} summarizes supported evidence for the instruction clauses and pairs defined above. Averaging S/$N$ within each eligible episode gives coverage of 85.6\%, 86.2\%, and 52.8\% for timing, order, and continuity, respectively, with each eligible episode receiving equal weight.

\begin{figure}[tbp]
\centering
\includegraphics[width=\columnwidth]{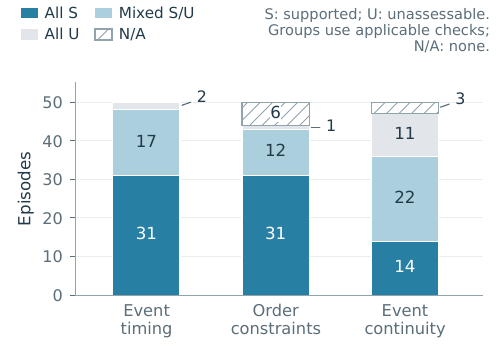}
\caption{\textbf{Temporal judgments with supplied history.} All 50 episodes are grouped by applicable checks: all supported (All S), mixed supported/unassessable (Mixed S/U), all unassessable (All U), or none (N/A). Counts measure evidence coverage under external context management, not unaided memory.}
\label{fig:temporal}
\end{figure}

\textbf{\textit{Distribution across episodes.}}
Figure~\ref{fig:temporal} complements the unit-level census by grouping episodes according to their applicable checks. All S means every applicable check has supported evidence; Mixed S/U combines supported and unassessable checks; All U means every applicable check is unassessable; and N/A denotes no applicable check.

Timing and order each have 31 All S episodes, compared with 14 for continuity. A further 17, 12, and 22 episodes, respectively, contain mixed evidence. Thus, the supported judgments extend across many episodes, while complete continuity coverage is less common. The six episodes without an applicable order pair and the three without an eligible continuity check remain visible as N/A, so every column retains the full cohort of 50.

In EP423, later views confirm an earlier turn's instruction match, without treating the confirmation as a new turn. In EP116, reviews distinguish the completed turn from the pending crossing, consistent with the lack of translation. These cases connect temporal judgments to the first and third findings: clarifying past actions and recognizing unfinished ones without completing them.

Unassessable checks include ambiguous boundaries, unexercised instruction events, and missing explicit later judgments. Continuity also has stricter eligibility and review requirements, so the group sizes do not rank temporal abilities. No contradiction was established in this preliminary audit; All U records insufficient evidence, not an observed temporal error. Supported responses show compatibility with recorded events under the supplied context. Without varying history selection or record retention, the counts do not separate model reasoning from the assistance provided by that context.

\subsection{Action Adjustment and Local Outcomes}
\label{sec:adaptive}
\textbf{Some viewpoint changes reveal previously hidden passages.}
We examine all 50 episodes using saved model proposals, RGB observations, and matching execution records. Proposals are interpreted alongside the workflow's movement constraints and selected actions. The preliminary offline, assistant-assisted audit identifies 66 problem-handling processes in 42 episodes; eight episodes contain no eligible process. Repeated attempts addressing the same problem are merged, while routine instruction following is excluded. Each process is grouped by its initial audited adjustment, and local relief requires evidence addressing the original difficulty, such as exposing an obscured passage or clearing an obstruction. These are observed system processes, not isolated tests of the model's planning contribution.

\begin{figure*}[t]
\centering
\includegraphics[width=\textwidth]{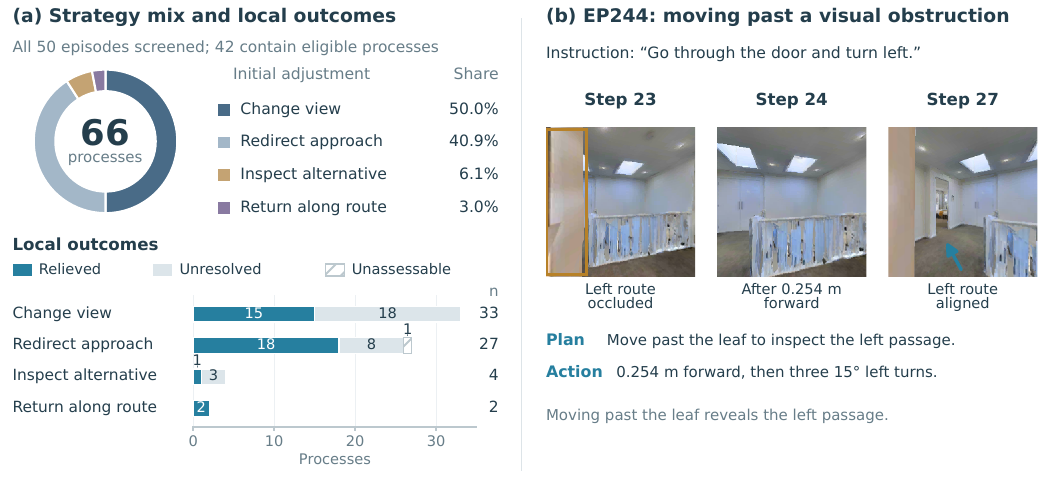}
\caption{\textbf{Action adjustments and local outcomes.} (a) The ring shows initial-strategy shares among 66 audited system processes; bars show local outcomes, with totals $n$ at the right. Hatching denotes unassessable outcomes. (b) EP244's executed viewpoint change reveals a passage hidden during stationary inspection. Gold outlines the door leaf; blue indicates the revealed passage. Plan text paraphrases the model proposal. Step~$t$ is the observation after $t$ actions.}
\label{fig:adaptive_census}
\end{figure*}

The ring in Figure~\ref{fig:adaptive_census}(a) summarizes the strategy mix: changes of viewing angle account for 33 processes (50.0\%), and adjustments to blocked approaches account for 27 (40.9\%). The remaining processes inspect another candidate (four, 6.1\%) or return along the earlier route (two, 3.0\%). The bars below show the corresponding local outcomes, including 15 relieved processes in the first group and 18 in the second. Across all groups, 36 processes show local relief, 29 remain unresolved at the observed end, and one is unassessable. These counts do not rank strategy effectiveness: a process may include several later adjustments, the groups encounter different problems, and the latter two groups are small.

Figure~\ref{fig:adaptive_census}(b) illustrates how an adjustment changes the available evidence. In EP244, stationary left/right inspections leave the landing's left continuation hidden by a nearby door leaf. At Step~23, GPT-6-Astra proposes a short forward movement to see past it. The recorded 0.254\,m translation exposes the left-side floor at Step~24; three left turns then align the view with the passage at Step~27. The useful outcome is the newly visible passage that answers the model's inspection question. This example illustrates the visual information seeking in the second finding; the process counts show both local relief and unresolved difficulties. Local relief remains distinct from final navigation success, and the recorded workflow does not establish a causal gain over continuing the previous strategy.

\subsection{Evidence-Guided Verification}
\label{sec:reflective}
\textbf{Some reviews resolve uncertainty through additional visual evidence.}
The workflow initiates progress and arrival reviews and supplies earlier records. GPT-6-Astra's recorded responses include questions, requests for saved images, and revised or retained judgments. Our preliminary assistant-assisted audit follows these verification processes across all 50 episodes, linking an earlier judgment, an attempt to obtain discriminating evidence, and a later judgment about the same reference or interpretation. Repeated reviews of an unresolved question form one process. Conclusions are checked against saved observations and execution records; the model's own support flag does not determine the audit label.

\begin{figure*}[t]
\centering
\includegraphics[width=\textwidth]{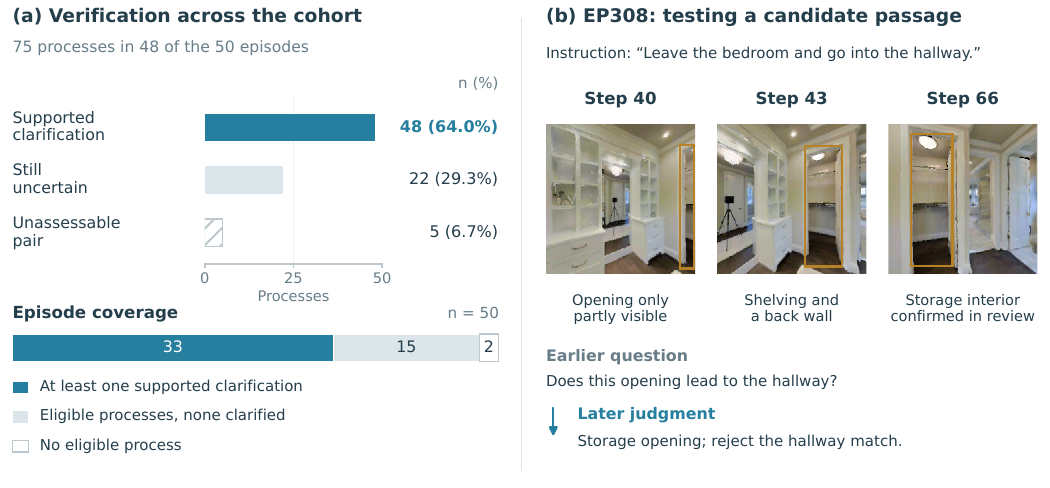}
\caption{\textbf{Evidence-guided verification in scheduled reviews.} (a) Outcomes of 75 processes and mutually exclusive coverage groups across all 50 episodes. Supported clarification includes justified candidate rejection. (b) EP308 resolves one opening as storage rather than the instructed hallway. Ochre outlines track the same candidate; judgment summaries paraphrase saved model reviews. Step $t$ denotes the observation after $t$ actions.}
\label{fig:reflective_census}
\end{figure*}

Figure~\ref{fig:reflective_census}(a) summarizes 75 verification processes in 48 episodes. Of these, 48 (64.0\%) resolve an initially uncertain judgment into an evidence-supported conclusion, including both identifying and justifiably rejecting a candidate. Another 22 (29.3\%) retain uncertainty, while five (6.7\%) have an unassessable endpoint. The episode bar counts each episode once: 33 contain at least one supported clarification, 15 contain eligible processes but no supported clarification, and two contain no eligible process. A local clarification does not imply that every question in that episode is resolved.

Figure~\ref{fig:reflective_census}(b) illustrates a supported rejection in EP308. At Step~40, GPT-6-Astra questions whether the partly visible opening leads to the instructed hallway. Further views expose shelving and a back wall; at Step~66, it explicitly rejects the opening as a hallway candidate. This resolves one competing interpretation while leaving the overall hallway route uncertain.

These processes show strengths in revising local judgments, while sometimes leaving the correct route unresolved. All independently assessable pairs begin with uncertainty, so this inventory provides no denominator for an error-correction rate. In the 22 unresolved processes, uncertainty persists through the observed reviews. These results describe judgment revision within workflow-scheduled calls; they do not establish spontaneous initiation of reflection or a causal benefit over a system without these reviews.

\section{Discussion and Limitations}
\label{sec:discussion}
\textbf{Later views support a match between an earlier action and the instruction.}
In EP423, later views allow GPT-6-Astra to confirm the earlier turn's role in the instructed route while retaining when it occurred. Across the evaluated episodes, all applicable timing and order checks have supported evidence in 31 episodes each, although timing checks alone do not establish a route match. The time of an action and the time when its instruction match becomes clear need not coincide. Assessing this distinction requires inspecting both the new observation and the supplied history, including the workflow's event records and order checks.

\textbf{Additional observations support local clarification in some cases.}
In EP244, a proposed forward movement exposes a hidden passage; in EP308, further views support rejecting an uncertain hallway candidate. Across all evaluated episodes, 36 of 66 adjustment processes show local relief and 48 of 75 verification processes produce a supported clarification. These findings suggest checking what an observation reveals, then whether that information changes navigation. Exposing a passage can be useful before advancing along it, whereas repeated reviews may leave the same question unresolved. The two inventories use different units and eligibility rules; their proportions do not measure a common success rate or establish causal navigation gains.

\textbf{The gap between understanding and completion spans movement and stopping.}
In EP116, GPT-6-Astra distinguishes a completed turn from an unfinished crossing while the system continues rotating. In EP308, rejecting one opening leaves the route unresolved. At termination, the 52.0\% endpoint success rate includes 16.0 percentage points from budget termination, while three accepted STOPs fail the distance criterion. These distinct outcomes share an evaluation requirement: follow the judgment through the selected action, actual motion, and stopping decision. The recorded sequence can locate where progress stops; identifying which model or controller choice caused the failure requires controlled comparisons.

\textbf{The results depend on both the model and the workflow.}
External code selects context, retains records, schedules reviews, and constrains actions. Requests receive selected earlier outputs without replaying full conversations or reasoning-state items. Trajectories measure system performance; responses show judgments under supplied context. Neither isolates unaided memory, spontaneous reflection, or intrinsic navigation ability. The data do not establish whether these workflow choices assist or restrict performance; retaining more context or removing a movement restriction is not demonstrated to help.

\textbf{Supported evidence is not a model accuracy score.}
The audits cover all 50 episodes but use different units: terminal references for grounding, instruction events for temporal judgments, and eligible adjustment or verification processes. Spatial analysis remains case-based. These measurements cannot be combined into a single capability score. Unassessable labels retain missing or ambiguous evidence rather than treating it as an error. The retrospective, assistant-assisted annotations have not undergone blinded human review or inter-rater validation, and model explanations are reported outputs rather than access to internal reasoning. Offline GT helps check route correspondence; it does not establish the outcome of an unexecuted alternative.

\textbf{Broader claims require fresh episodes and matched comparisons.}
The evaluated subset is a fixed 50-episode prefix spanning nine scenes, with one recorded run per episode and exposure during workflow development. It does not estimate run-to-run variability or full-split performance. Published results with different cohorts, data versions, observation access, budgets, and stopping rules provide context without a controlled superiority comparison. Future evaluation should freeze these conditions, use fresh episodes and repeated runs, and independently validate the annotations. Holding the workflow fixed while substituting models would compare their contributions within that workflow; holding the model fixed while varying context retention would test the effect of context management. Following clarified judgments through subsequent actions would further test whether local resolution persists into route progress and appropriate stopping.

\section{Conclusion}
\label{sec:conclusion}
The evaluated GPT-6-Astra-based navigation system achieves 52.0\% SR, 48.9\% SPL, and 70.8\% nDTW with external context management and action control. Its recorded responses show strengths in linking landmarks and earlier actions to instructions, seeking visual information, and revising uncertain judgments. The observed gap between task understanding and autonomous completion includes continued rotation after an unfinished crossing is recognized. At termination, eighteen episodes succeed with a workflow-accepted STOP, eight succeed at the budget limit, and three accepted STOPs fail the distance criterion. These findings concern the complete system and make the next challenge concrete: translating supported judgments into the required movement, sustained route progress, and appropriate stopping.

{\small
\bibliographystyle{ieeenat_fullname}
\bibliography{main}

@String(CVPR= {IEEE Conf. Comput. Vis. Pattern Recog.})

@String(ICCV= {Int. Conf. Comput. Vis.})

@String(ECCV= {Eur. Conf. Comput. Vis.})

@String(AAAI = {AAAI})

@String(CVPR  = {CVPR})

@String(ICCV  = {ICCV})

@String(ECCV  = {ECCV})

@inproceedings{r2r,
	author    = {Peter Anderson and
	Qi Wu and
	Damien Teney and
	Jake Bruce and
	Mark Johnson and
	Niko S{\"{u}}nderhauf and
	Ian D. Reid and
	Stephen Gould and
	Anton van den Hengel},
	title     = {Vision-and-Language Navigation: Interpreting Visually-Grounded Navigation
	Instructions in Real Environments},
	booktitle = CVPR,
	pages     = {3674--3683},
	year      = {2018},
}

@inproceedings{vlnce,
  author    = {Jacob Krantz and
               Erik Wijmans and
               Arjun Majumdar and
               Dhruv Batra and
               Stefan Lee},
  editor    = {Andrea Vedaldi and
               Horst Bischof and
               Thomas Brox and
               Jan{-}Michael Frahm},
  title     = {Beyond the Nav-Graph: Vision-and-Language Navigation in Continuous
               Environments},
  booktitle = ECCV,
  pages     = {104--120},
  year      = {2020}
}

@InProceedings{dcvln,
    author    = {Hong, Yicong and Wang, Zun and Wu, Qi and Gould, Stephen},
    title     = {Bridging the Gap Between Learning in Discrete and Continuous Environments for Vision-and-Language Navigation},
    booktitle = CVPR,
    pages     = {15418--15428},
    year      = {2022}
}

@inproceedings{an2023bevbert,
  title={Bevbert: Multimodal map pre-training for language-guided navigation},
  author={An, Dong and Qi, Yuankai and Li, Yangguang and Huang, Yan and Wang, Liang and Tan, Tieniu and Shao, Jing},
  booktitle=ICCV,
  pages={2737--2748},
  year={2023}
}

@inproceedings{wang2023scaling,
  title={Scaling data generation in vision-and-language navigation},
  author={Wang, Zun and Li, Jialu and Hong, Yicong and Wang, Yi and Wu, Qi and Bansal, Mohit and Gould, Stephen and Tan, Hao and Qiao, Yu},
  booktitle=ICCV,
  pages={12009--12020},
  year={2023}
}

@inproceedings{qiao2025opennav,
  author    = {Yanyuan Qiao and Wenqi Lyu and Hui Wang and Zixu Wang and Zerui Li and Yuan Zhang and Mingkui Tan and Qi Wu},
  title     = {Open-Nav: Exploring Zero-Shot Vision-and-Language Navigation in Continuous Environment with Open-Source LLMs},
  booktitle = {ICRA},
  year      = {2025},
  pages={6710-6717}
}

@inproceedings{shi2025smartway,
  title={Smartway: Enhanced waypoint prediction and backtracking for zero-shot vision-and-language navigation},
  author={Shi, Xiangyu and Li, Zerui and Lyu, Wenqi and Xia, Jiatong and Dayoub, Feras and Qiao, Yanyuan and Wu, Qi},
  booktitle={IROS},
  pages={16923--16930},
  year={2025}
}

@article{shi2025fastsmartway,
  title={Fast-SmartWay: Panoramic-Free End-to-End Zero-Shot Vision-and-Language Navigation},
  author={Shi, Xiangyu and Li, Zerui and Qiao, Yanyuan and Wu, Qi},
  journal={arXiv preprint arXiv:2511.00933},
  year={2025}
}

@article{qiao2025navbench,
  title={NavBench: Probing Multimodal Large Language Models for Embodied Navigation},
  author={Qiao, Yanyuan and Hong, Haodong and Lyu, Wenqi and An, Dong and Zhang, Siqi and Xie, Yutong and Wang, Xinyu and Wu, Qi},
  journal={arXiv preprint arXiv:2506.01031},
  year={2025}
}

@article{wang2025dreamnav,
  title={DreamNav: A Trajectory-Based Imaginative Framework for Zero-Shot Vision-and-Language Navigation},
  author={Wang, Yunheng and Fang, Yuetong and Wang, Taowen and Feng, Yixiao and Tan, Yawen and Zhang, Shuning and Liu, Peiran and Ji, Yiding and Xu, Renjing},
  journal={arXiv preprint arXiv:2509.11197},
  year={2025}
}

@article{liu2024energy,
  title={Vision-language navigation with energy-based policy},
  author={Liu, Rui and Wang, Wenguan and Yang, Yi},
  journal={Advances in Neural Information Processing Systems},
  volume={37},
  pages={108208--108230},
  year={2024}
}

@misc{vlnceDataV13,
  author = {{VLN-CE Project}},
  title = {{VLN-CE} Data: {R2R\_VLNCE\_v1-3}},
  year = {2022},
  howpublished = {\url{https://jacobkrantz.github.io/vlnce/data}},
  note = {Dataset release notes; accessed September 15, 2026}
}

@inproceedings{zhou2024navgpt_verified,
  author = {Zhou, Gengze and Hong, Yicong and Wu, Qi},
  title = {{NavGPT}: Explicit Reasoning in Vision-and-Language Navigation with Large Language Models},
  booktitle = {Proceedings of the AAAI Conference on Artificial Intelligence},
  volume = {38},
  number = {7},
  pages = {7641--7649},
  year = {2024},
  doi = {10.1609/aaai.v38i7.28597},
  url = {https://ojs.aaai.org/index.php/AAAI/article/view/28597}
}

@inproceedings{long2024discussnav_verified,
  author = {Long, Yuxing and Li, Xiaoqi and Cai, Wenzhe and Dong, Hao},
  title = {Discuss Before Moving: Visual Language Navigation via Multi-expert Discussions},
  booktitle = {2024 IEEE International Conference on Robotics and Automation (ICRA)},
  year = {2024},
  url = {https://arxiv.org/abs/2309.11382}
}

@inproceedings{chen2024mapgpt_verified,
  author = {Chen, Jiaqi and Lin, Bingqian and Xu, Ran and Chai, Zhenhua and Liang, Xiaodan and Wong, Kwan-Yee},
  title = {{MapGPT}: Map-Guided Prompting with Adaptive Path Planning for Vision-and-Language Navigation},
  booktitle = {Proceedings of the 62nd Annual Meeting of the Association for Computational Linguistics (Volume 1: Long Papers)},
  pages = {9796--9810},
  year = {2024},
  doi = {10.18653/v1/2024.acl-long.529},
  url = {https://aclanthology.org/2024.acl-long.529/}
}

@inproceedings{dai2026evonav,
  author = {Dai, Guangzhao and Wang, Shuo and Wang, Zihan and Xie, Guo-Sen and Yang, Yang and Pan, Jinshan and Sun, Qianru and Shu, Xiangbo},
  title = {History to Future: Evolving Agent with Experience and Thought for Zero-shot Vision-and-Language Navigation},
  booktitle = {Proceedings of the IEEE/CVF Conference on Computer Vision and Pattern Recognition (CVPR)},
  month = jun,
  year = {2026},
  pages = {15177--15187},
  url = {https://openaccess.thecvf.com/content/CVPR2026/html/Dai_History_to_Future_Evolving_Agent_with_Experience_and_Thought_for_CVPR_2026_paper.html}
}

@inproceedings{xue2026profocus,
  author = {Xue, Wei and Li, Mingcheng and Wu, Xuecheng and Tang, Jingqun and Yang, Dingkang and Zhang, Lihua},
  title = {{ProFocus}: Proactive Perception and Focused Reasoning in Vision-and-Language Navigation},
  booktitle = {Proceedings of the IEEE/CVF Conference on Computer Vision and Pattern Recognition (CVPR)},
  month = jun,
  year = {2026},
  pages = {18129--18139},
  url = {https://openaccess.thecvf.com/content/CVPR2026/html/Xue_ProFocus_Proactive_Perception_and_Focused_Reasoning_in_Vision-and-Language_Navigation_CVPR_2026_paper.html}
}

@article{li2026agenticnav,
  author = {Li, Yijian and Li, Changze and Shi, Hantian and Luo, Jiaying and Cai, Jiyuan and Yang, Ming and Qin, Tong},
  title = {{AgenticNav}: Zero-Shot Vision-and-Language Navigation as a Tool-Calling Harness},
  journal = {arXiv preprint arXiv:2606.10577},
  year = {2026},
  doi = {10.48550/arXiv.2606.10577},
  url = {https://arxiv.org/abs/2606.10577}
}

@inproceedings{ilharco2019ndtw_verified,
  author = {Ilharco, Gabriel and Jain, Vihan and Ku, Alexander and Ie, Eugene and Baldridge, Jason},
  title = {General Evaluation for Instruction Conditioned Navigation using Dynamic Time Warping},
  booktitle = {NeurIPS Workshop on Visually Grounded Interaction and Language (ViGIL)},
  year = {2019},
  url = {https://vigilworkshop.github.io/static/papers-2019/33.pdf}
}

@article{an2025etpnav_verified,
  author = {An, Dong and Wang, Hanqing and Wang, Wenguan and Wang, Zun and Huang, Yan and He, Keji and Wang, Liang},
  title = {{ETPNav}: Evolving Topological Planning for Vision-Language Navigation in Continuous Environments},
  journal = {IEEE Transactions on Pattern Analysis and Machine Intelligence},
  volume = {47},
  number = {7},
  pages = {5130--5145},
  year = {2025},
  doi = {10.1109/TPAMI.2024.3386695},
  url = {https://ieeexplore.ieee.org/document/10495141/}
}

@article{chen2025canav_verified,
  author = {Chen, Kehan and An, Dong and Huang, Yan and Xu, Rongtao and Su, Yifei and Ling, Yonggen and Reid, Ian and Wang, Liang},
  title = {{Constraint-Aware Zero-Shot Vision-Language Navigation in Continuous Environments}},
  year = {2025},
  url = {https://arxiv.org/abs/2412.10137},
  journal = {IEEE Transactions on Pattern Analysis and Machine Intelligence},
  volume = {47},
  number = {11},
  pages = {10441--10456}
}

@inproceedings{chen2025aoplanner_verified,
  author = {Jiaqi Chen and Bingqian Lin and Xinmin Liu and Lin Ma and Xiaodan Liang and Kwan-Yee K. Wong},
  title = {{Affordances-Oriented Planning Using Foundation Models for Continuous Vision-Language Navigation}},
  year = {2025},
  url = {https://ojs.aaai.org/index.php/AAAI/article/view/34526},
  booktitle = {Proceedings of the AAAI Conference on Artificial Intelligence},
  doi = {10.1609/aaai.v39i22.34526},
  pages = {23568--23576}
}

@inproceedings{wang2025g3dlf_verified,
  author = {Wang, Zihan and Lee, Gim Hee},
  title = {{g3D-LF: Generalizable 3D-Language Feature Fields for Embodied Tasks}},
  year = {2025},
  url = {https://openaccess.thecvf.com/content/CVPR2025/html/Wang_g3D-LF_Generalizable_3D-Language_Feature_Fields_for_Embodied_Tasks_CVPR_2025_paper.html},
  booktitle = {CVPR},
  pages = {14191--14202}
}

@inproceedings{zhang2024navid_rss,
  author = {Zhang, Jiazhao and Wang, Kunyu and Xu, Rongtao and Zhou, Gengze and Hong, Yicong and Fang, Xiaomeng and Wu, Qi and Zhang, Zhizheng and Wang, He},
  title = {{NaVid: Video-based VLM Plans the Next Step for Vision-and-Language Navigation}},
  year = {2024},
  url = {https://arxiv.org/abs/2402.15852},
  booktitle = {Robotics: Science and Systems}
}

@inproceedings{zhang2025uninavid_rss,
  author = {Jiazhao Zhang and Kunyu Wang and Shaoan Wang and Minghan Li and Haoran Liu and Songlin Wei and Zhongyuan Wang and Zhizheng Zhang and He Wang},
  title = {{Uni-NaVid: A Video-based Vision-Language-Action Model for Unifying Embodied Navigation Tasks}},
  year = {2025},
  url = {https://www.roboticsproceedings.org/rss21/p013.html},
  booktitle = {Robotics: Science and Systems},
  doi = {10.15607/RSS.2025.XXI.013}
}

@inproceedings{cheng2025navila_rss,
  author = {An-Chieh Cheng and Yandong Ji and Zhaojing Yang and Zaitian Gongye and Xueyan Zou and Jan Kautz and Erdem Biyik and Hongxu Yin and Sifei Liu and Xiaolong Wang},
  title = {{NaVILA: Legged Robot Vision-Language-Action Model for Navigation}},
  year = {2025},
  url = {https://www.roboticsproceedings.org/rss21/p018.html},
  booktitle = {Robotics: Science and Systems},
  doi = {10.15607/RSS.2025.XXI.018}
}

@inproceedings{wang2025auxthink_neurips,
  author = {Wang, Shuo and Wang, Yongcai and Li, Wanting and Cai, Xudong and Wang, Yucheng and Chen, Maiyue and Wang, Kaihui and Su, Zhizhong and Li, Deying and Fan, Zhaoxin},
  title = {{Aux-Think: Exploring Reasoning Strategies for Data-Efficient Vision-Language Navigation}},
  year = {2025},
  url = {https://proceedings.neurips.cc/paper_files/paper/2025/hash/2c90bf5bffe6497730ecade3a3a37458-Abstract-Conference.html},
  booktitle = {Advances in Neural Information Processing Systems},
  volume = {38}
}

@inproceedings{wang2025dynam3d_neurips,
  author = {Wang, Zihan and Lee, Seungjun and Lee, Gim Hee},
  title = {{Dynam3D: Dynamic Layered 3D Tokens Empower VLM for Vision-and-Language Navigation}},
  year = {2025},
  url = {https://proceedings.neurips.cc/paper_files/paper/2025/hash/e1a1847e39a7b79b41199176b152f0e6-Abstract-Conference.html},
  booktitle = {Advances in Neural Information Processing Systems},
  volume = {38}
}

@inproceedings{wei2026streamvln_icra,
  author = {Wei, Meng and Wan, Chenyang and Yu, Xiqian and Wang, Tai and Yang, Yuqiang and Mao, Xiaohan and Zhu, Chenming and Cai, Wenzhe and Wang, Hanqing and Chen, Yilun and Liu, Xihui and Pang, Jiangmiao},
  title = {{StreamVLN: Streaming Vision-and-Language Navigation via SlowFast Context Modeling}},
  year = {2026},
  url = {https://arxiv.org/abs/2507.05240},
  booktitle = {IEEE International Conference on Robotics and Automation},
  note = {ICRA 2026 acceptance stated in the authors' arXiv record}
}

@inproceedings{wang2026d3dvlp_verified,
  author = {Zihan Wang and Seungjun Lee and Guangzhao Dai and Gim Hee Lee},
  title = {{D3D-VLP: Dynamic 3D Vision-Language-Planning Model for Embodied Grounding and Navigation}},
  year = {2026},
  url = {https://openaccess.thecvf.com/content/CVPR2026/html/Wang_D3D-VLP_Dynamic_3D_Vision-Language-Planning_Model_for_Embodied_Grounding_and_Navigation_CVPR_2026_paper.html},
  booktitle = {CVPR},
  pages = {32463--32474}
}

@inproceedings{yin2025gcvln_corl,
  author = {Yin, Hang and Wei, Haoyu and Xu, Xiuwei and Guo, Wenxuan and Zhou, Jie and Lu, Jiwen},
  title = {{GC-VLN: Instruction as Graph Constraints for Training-free Vision-and-Language Navigation}},
  year = {2025},
  url = {https://proceedings.mlr.press/v305/yin25a.html},
  booktitle = {Proceedings of The 9th Conference on Robot Learning},
  pages = {1809--1824},
  volume = {305}
}

@inproceedings{li2026hsgm_verified,
  author = {Li, Kailing and Qian, Tianwen and Yang, Lijin and Fu, Yuqian and Gong, Jingyu and Wang, Xiaoling and He, Liang},
  title = {{Bridging the 2D-3D Gap: A Hierarchical Semantic-Geometric Map for Vision Language Navigation}},
  year = {2026},
  url = {https://arxiv.org/abs/2606.00095},
  booktitle = {CVPR}
}

@article{chen2026harnessvln_verified,
  author = {Chen, Yang and Che, Lirong and Huang, Zhenyu and Fu, Wenbo and Wang, Chuang and Cao, Xu and Liu, Daqi and Yang, Yuzhe and Su, Jian and Guo, Lan-Zhe},
  title = {{HarnessVLN: Unifying Training-Free Embodied Navigation through an Agent Harness}},
  year = {2026},
  url = {https://arxiv.org/abs/2609.15195v1},
  journal = {arXiv preprint arXiv:2609.15195}
}

@article{zhang2026spatialant_verified,
  author = {Zhang, Jiwen and Shi, Xiangyu and Wang, Siyuan and Li, Zerui and Wei, Zhongyu and Wu, Qi},
  title = {{SpatialAnt: Autonomous Zero-Shot Robot Navigation via Active Scene Reconstruction and Visual Anticipation}},
  year = {2026},
  url = {https://arxiv.org/abs/2603.26837},
  journal = {arXiv preprint arXiv:2603.26837}
}

@inproceedings{wang2024hnr_verified,
  author = {Wang, Zihan and Li, Xiangyang and Yang, Jiahao and Liu, Yeqi and Hu, Junjie and Jiang, Ming and Jiang, Shuqiang},
  title = {{Lookahead Exploration with Neural Radiance Representation for Continuous Vision-Language Navigation}},
  booktitle = {Proceedings of the IEEE/CVF Conference on Computer Vision and Pattern Recognition},
  year = {2024},
  month = {June},
  pages = {13753--13762},
  url = {https://openaccess.thecvf.com/content/CVPR2024/html/Wang_Lookahead_Exploration_with_Neural_Radiance_Representation_for_Continuous_Vision-Language_Navigation_CVPR_2024_paper.html}
}

@inproceedings{ding2026lavira_icra,
  author = {Ding, Hongyu and Xu, Ziming and Fang, Yudong and Wu, You and Chen, Zixuan and Shi, Jieqi and Huo, Jing and Zhang, Yifan and Gao, Yang},
  title = {{LaViRA: Language-Vision-Robot Actions Translation for Zero-Shot Vision Language Navigation in Continuous Environments}},
  booktitle = {IEEE International Conference on Robotics and Automation},
  year = {2026},
  url = {https://arxiv.org/abs/2510.19655},
  eprint = {2510.19655},
  archivePrefix = {arXiv},
  primaryClass = {cs.RO},
  note = {ICRA 2026 acceptance stated in the authors' arXiv record; results from version 2}
}

@inproceedings{zheng2026threestepnav,
  author = {Zheng, Wanrong and Ge, Yunhao and Itti, Laurent},
  title = {{Three-Step Nav: A Hierarchical Global--Local Planner for Zero-Shot Vision-and-Language Navigation}},
  booktitle = {Proceedings of The 29th International Conference on Artificial Intelligence and Statistics},
  series = {Proceedings of Machine Learning Research},
  volume = {300},
  pages = {4645--4653},
  publisher = {PMLR},
  year = {2026},
  url = {https://proceedings.mlr.press/v300/zheng26b.html}
}

@article{zheng2026c2nav,
  author = {Zheng, Runtian and Zhang, Congpeng and Liu, Ying},
  title = {{C$^{2}$Nav: Compare Before You Commit for Zero-Shot Vision-and-Language Navigation}},
  journal = {arXiv preprint arXiv:2609.15142},
  year = {2026},
  url = {https://arxiv.org/abs/2609.15142}
}
}
\end{document}